\pdfoutput=1

\documentclass[11pt]{article}

\usepackage[final]{acl}

\usepackage{times}
\usepackage{latexsym}

\usepackage[T1]{fontenc}

\usepackage[utf8]{inputenc}

\usepackage{microtype}

\usepackage{multirow}

\usepackage{inconsolata}

\usepackage{graphicx}

\usepackage{amsmath}

\usepackage{float}

\usepackage{array}

\usepackage{fancyhdr}
\fancypagestyle{firstpage}{
  \fancyhf{}                    
  \fancyhead[L]{\normalsize Accepted at ACL 2025} 
  
}
\title{SpeechWeave: Diverse Multilingual Synthetic Text \& Audio Data Generation Pipeline for Training Text to Speech Models}

 \author{Karan Dua, Puneet Mittal,
  Ranjeet Gupta,
  Hitesh Laxmichand Patel \\
  \{karan.dua, puneet.mittal, ranjeet.gupta, hitesh.laxmichand.patel\}@oracle.com \\\\
         Oracle AI}

\begin{document}
    \maketitle
    \thispagestyle{firstpage}
\begin{abstract}
High-quality Text-to-Speech (TTS) model training requires extensive and diverse text and speech data. It is challenging to procure such data from real sources due to issues of domain specificity, licensing, and scalability. Large language models (LLMs) can certainly generate textual data, but they create repetitive text with insufficient variation in the prompt during the generation process. Another important aspect in TTS training data is text normalization. Tools for normalization might occasionally introduce anomalies or overlook valuable patterns, and thus impact data quality. Furthermore, it is also impractical to rely on voice artists for large scale speech recording in commercial TTS systems with standardized voices. To address these challenges, we propose \textbf{SpeechWeave}, a synthetic speech data generation pipeline that is capable of automating the generation of multilingual, domain-specific datasets for training TTS models. Our experiments reveal that our pipeline generates data that is \textbf{10–48\%} more diverse than the baseline across various linguistic and phonetic metrics, along with speaker-standardized speech audio while generating approximately \textbf{97\%} correctly normalized text. Our approach enables scalable, high-quality data generation for TTS training, improving diversity, normalization, and voice consistency in the generated datasets.
\end{abstract}

\section{Introduction}
Text-to-Speech (TTS) systems convert written text to spoken audio and are used in applications such as virtual assistants, accessibility software, navigation systems, and customer service to enable easier and accessible user interaction. TTS systems require massive amounts of training data consisting of text and speech pairs. Most publicly available TTS datasets include book readings or generic passages \cite{ljspeech17}, \cite{7178964},, \cite{ardila2020commonvoicemassivelymultilingualspeech}. However, for domain-specific business data (e.g., Automobile, Healthcare, Retail), one needs to either scrape it from the web or purchase it from data curation companies, which could introduce cost and licensing issues. Additionally, the multilingual nature of such systems complicates the process of obtaining domain-specific data.
\begin{figure}[h!]
  \includegraphics[width=1.0\linewidth]{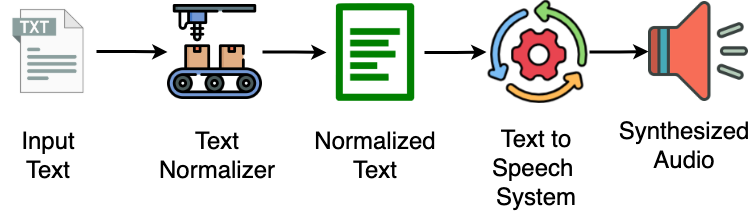}
  \caption {\label{fig:tts-pipeline}High-level depiction of a TTS system incorporating normalization}
\end{figure}
\begin{table*}[h!]
  \centering
  \small
  \renewcommand{\arraystretch}{1.3}
  \begin{tabular}{>{\centering\arraybackslash}p{2cm}p{1cm}p{11.5cm}}
    \hline
    \textbf{Temperature} & \textbf{top\_p} & \textbf{Output (Generated 3 times with the same settings)} \\
    \hline
    0.7 & 0.95 & 1. Maintaining a budget surplus is essential for reducing debt and securing financial stability. \\
        &      & 2. Maintaining a budget surplus can help reduce debt and ensure financial stability. \\
        &      & 3. Maintaining a balanced budget is crucial for achieving financial stability. \\
    \hline
    1.2 & 0.95 & 1. Managing expenses effectively is crucial for maintaining a healthy financial situation. \\
        &      & 2. Making regular investments in a diversified portfolio can help grow your wealth over time. \\
        &      & 3. Maintaining a balanced budget is key to achieving long-term financial stability. \\  
    \hline
    1.2 & 0.51 & 1. Maintaining a budget surplus is crucial for reducing debt and ensuring financial stability. \\
        &      & 2. Maintaining a budget helps control your expenses and save for future financial goals. \\
    \hline
    2.5 & 0.95 & 1. A wiser investment would have made twice your savings in a safer market stance. (Implying the reader missed an opportunity for increased returns by prioritizing safeguarded funds) \\  
        &      & 2. Exporting Savings: Despite decreased expenses, investments yield higher returns, bolstering my financial security. \\  
        &      & 3. A sound financial plan includes creating a budget, building an emergency reserves, saving for retirement, and consistently reducing unnecessary debts. But if you intend a single short sentence it would be- Investing as early and aggressively thanYou Can can enhance long-term returns significantly due toCompainddffects.   \\
    \hline
  \end{tabular}
  \caption{\label{tab:prompting-exps}Generated outputs for different temperature and top-p settings by prompting an LLM directly. Prompt used was "Construct one short sentence in the finance domain".}
\end{table*}

\vspace{-2mm}
\subsection{Text Scripts}
Text scripts serve as inputs to TTS systems and are essential for adapting these systems to specific business domains. While obtaining domain-specific data from public sources is difficult, it may seem that LLMs can easily generate the necessary data using a simple prompt with the domain as input. However, our experiments with \emph{Mistral-7b-Instruct} \cite{jiang2023mistral7b} show that for short sentences, the generated text remains similar even with high \emph{temperature} and \emph{top\_p} values, especially if the input prompt stays unchanged. As shown in Table~\ref{tab:prompting-exps}, an LLM, even high temperature values produce almost identical results. Very high values still limit the sub-domain to \emph{Personal Finance} but may also generate unstable, low-quality output. Our analysis in the~\nameref{sec:diversity-analysis} section shows that LLMs, without prompt variation, result in low-diversity datasets, thus making this approach impractical for generating large datasets for training downstream models.

\subsection{Normalization}
The written and spoken forms of text often differ, primarily in specific entities like addresses, dates, times, and salutations, known as semiotic classes \cite{10.1162/coli_a_00349}. Table~\ref{tab:normalization-examples} presents examples of text scripts with their normalized forms across languages.
\begin{table}[h!]
  \centering
  \small
  \renewcommand{\arraystretch}{1.3}
  \begin{tabular}{p{1cm} p{2.5cm} p{2.5cm}}
    \hline
    \multicolumn{1}{c}{\textbf{Language}} & \multicolumn{1}{c}{\textbf{Text Script}} & \multicolumn{1}{c}{\textbf{Normalized Form}} \\
    \hline
    English & The best waffles in Delhi are found in the \textbf{10th St., Hauz Khas Vil.} in South Delhi. & The best waffles in Delhi are found in the \textbf{tenth street, Hauz Khas Village} in South Delhi. \\
    Spanish & \textbf{El Dr.} Johnson se especializa en el manejo de enfermedades relacionadas con el estilo de vida. & \textbf{El Doctor} Johnson se especializa en el manejo de enfermedades relacionadas con el estilo de vida. \\
    French & Emily est née le \textbf{03/08/1995}. & Emily est née le \textbf{trois août mil neuf cent quatre-vingt-quinze.} \\
    \hline
  \end{tabular}
  \caption{\label{tab:normalization-examples} Examples of text scripts along with their normalized forms across semiotic classes and languages.}
\end{table}
A TTS system processes text through a Text Normalization System, such as NeMo's text normalizer \cite{zhang2021nemoinversetextnormalization}, before generating speech audio, as depicted in Figure~\ref{fig:tts-pipeline}. However, normalization systems may have limitations, failing to recognize all semiotic class variations. For example, a date could appear as 03/01/2005, 01-Mar-2005, or March 01, 2005, and some formats may be overlooked.
For inference, a pre-processing text normalizer is essential. However, for training data generation, our work demonstrates that normalizing semiotic classes at the time of generation achieves higher accuracy, eliminating the need for a separate text normalizer.
\begin{figure*}[h!]
  \includegraphics[width=1.0\linewidth]{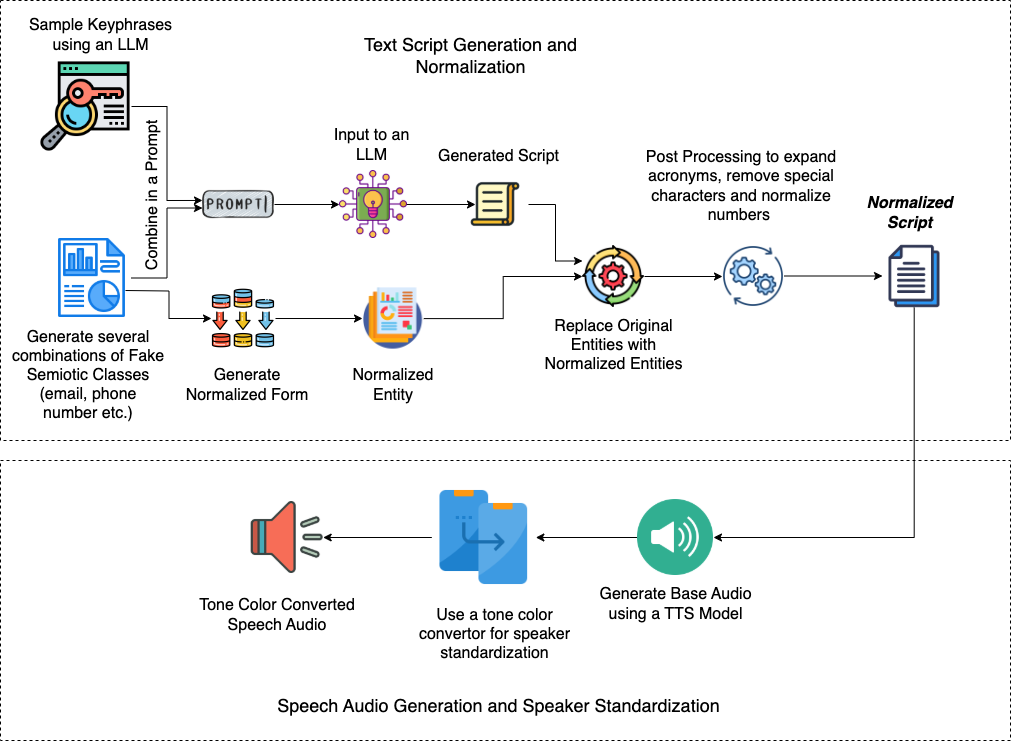}
  \caption {\label{fig:high-level-pipeline}High-level description of our synthetic text and audio generation pipeline}
\end{figure*}
\subsection{Audio Data}
Commercial TTS systems require speaker standardization to allow customers to choose a specific speaker based on their usecase. To achieve this, TTS systems need training data tailored to these specific speakers. Utilizing human voice artists to record speech audio for curating such training data is expensive and therefore not scalable.
\\
\\
To address these challenges, we introduce \textbf{SpeechWeave}—\textit{a comprehensive synthetic speech data generation pipeline}. Our key contributions through SpeechWeave include:  

\begin{itemize}
    \item An end-to-end automated pipeline for generating high-quality synthetic data to train Text-to-Speech models.
    \item Highly diverse text generation—both linguistically and phonetically—with thousands of unique combinations of semiotic classes, normalized at the source with high accuracy.
    \item High-quality speech audio generation with speaker standardization to ensure consistency in speech characteristics for commercial TTS systems.
\end{itemize}

\section{Related Work}
\cite{holtzman2020curiouscaseneuraltext} introduced nucleus sampling to stabilize text diversity in language models. Studies like \cite{naik2024diversitythoughtimprovesreasoning} and \cite{li2023makinglargelanguagemodels} explored prompt engineering techniques to improve LLM performance. \cite{meincke2024promptingdiverseideasincreasing} highlighted LLM limitations in generating diverse ideas, showing how strategies like Chain-of-Thought prompting can help. \cite{hayati2024farextractdiverseperspectives} focused on step-by-step recall prompting for diversity.

\cite{cornell2024generatingdatatexttospeechlargelanguage} proposed a pipeline combining LLM-generated text and TTS for ASR data, while Gunduz et al. \cite{gunduz-etal-2024-automated} introduced an open-source TTS data generation tool that lacked text script generation and normalization, relying on public datasets like the Opus corpus \cite{4992de1b5fb34f3e9691772606b36edf} and voice artists for recordings. \cite{hsu2024lowresourceselfsupervisedlearningsslenhanced} presented a low-resource, self-supervised method for training TTS using unlabeled audio.

Works like \cite{eldan2023tinystoriessmalllanguagemodels} and \cite{cox2023promptinglargelanguagemodel} showed how keyphrases can increase text diversity in LLMs. In TTS, \cite{9689505} trained a multi-speaker model for low-resource languages, while \cite{qin2024openvoiceversatileinstantvoice} developed a cross-lingual tone converter for vocal characteristics.

Other studies, like \cite{zhang2021nemoinversetextnormalization}, \cite{mansfield-etal-2019-neural}, and \cite{ro2022transformerbasedmodelstextnormalization}, focused on text normalization systems.

Despite these advancements, no prior work has proposed an integrated pipeline for generating diverse text scripts and their normalized forms and speaker standardized speech audio for TTS training.

\section{Our Pipeline and Components}
\textbf{SpeechWeave} consists of a \textbf{keyphrase sampler}, an \textbf{entity sampler with at-source normalizer}, a \textbf{postprocessor} and an \textbf{audio generation} module.

The pipeline is depicted at a high level in Figure~\ref{fig:high-level-pipeline}. A more detailed representation is available in Figure~\ref{fig:text-script-pipeline} in Appendix.

\subsection{Keyphrase Sampling}
As noticed above, if there isn't enough diversity in the inputs to an LLM, the model tends to generate repetitive text. One way to improve the diversity of generated text is through keyphrase infusion in prompts as demonstrated by \cite{eldan2023tinystoriessmalllanguagemodels}. For e.g. instead of prompting the model "Generate a sentence in finance domain", we can prompt, "Generate a sentence in finance domain containing the following keyphrases: Mortgage, Asset Finance". We can prompt the model to generate text with multiple such keyphrase combinations to ensure higher diversity in the generated text. 
\subsubsection{Multi-Step Prompting}
For domain-specific keyphrases, we may prompt an LLM to generate them, but this can lead to repetition. To address this, we use a multi-step prompting approach. As shown by \cite{hayati2024farextractdiverseperspectives}, iterative multi-step prompting enhances idea diversity. We begin by generating a list of subdomains within a business domain, such as healthcare. Then we randomly select one from the generated list. The LLM is then prompted to generate a creative paragraph for the chosen subdomain, and then we prompt the LLM to extract relevant keyphrases. To ensure structured output, we use lm-format-enforcer \cite{Gat} to convert results into a parseable JSON format at each step.
\begin{table*}[t]
  \centering
  \small
  \renewcommand{\arraystretch}{1.3}
  \begin{tabular}{p{1.5cm} p{3.5cm} p{1.4cm} p{1.4cm} p{1.5cm} p{1.0cm} p{1.0cm} lp{1.2cm}}
    \hline
    \textbf{Language} & \textbf{Dataset} & \textbf{\begin{tabular}[c]{@{}l@{}}Mean \\ Similarity \\Score \\ (Grouped)\end{tabular}} & \textbf{\begin{tabular}[c]{@{}l@{}}Max \\Similarity \\ Score \\ (Grouped)\end{tabular}} & \textbf{\begin{tabular}[c]{@{}l@{}}Mean \\Similarity \\ Score \\ (Ungrouped)\end{tabular}} & \textbf{TTR} & \textbf{MATTR}  & \textbf{\begin{tabular}[c]{@{}l@{}}Diphone \\ Coverage\end{tabular}}  \\
    \hline
    \multirow{3}{*}{English} 
    & Direct Prompting (Baseline) & 0.48 & 0.70 & 0.22 & 0.118 & 0.761& 1442  \\
    & English LibriSpeech & - & - & 0.36 & 0.123 & 0.758 & \textbf{1792}  \\
    & Ours & \textbf{0.26} & \textbf{0.36} & \textbf{0.15} & \textbf{0.167} & \textbf{0.803}  & 1694  \\
    \hline
    \multirow{3}{*}{Spanish} 
    & Direct Prompting (Baseline) & 0.54 & 0.77 & 0.31 & 0.297 & 0.966 & 516  \\
    & Spanish LibriSpeech & - & - & 0.28 & \textbf{0.395} & 0.962 & \textbf{651}  \\
    & Ours & \textbf{0.30} & \textbf{0.41} & \textbf{0.25}  & 0.370 & \textbf{0.979} & 565  \\
    \hline
  \end{tabular}
  \caption{Comparison of similarity scores, lexical diversity (TTR, MATTR), and phoneme coverage (Diphone Coverage) between our method, direct prompting baseline, and public datasets.}
  \label{tab:merged-comparison}
\end{table*}
\subsubsection{Keyphrase Store and De-Duplication}
We utilize an in-memory keyphrase store to store domain and language specific keyphrases. We also utilize fuzzy search based on token sort ratio and Levenshtein distance to ensure that we do not store keyphrases that are very similar to each other. This can also be replaced with a keyphrase embeddings model such as PhraseBERT \cite{wang2021phrasebertimprovedphraseembeddings}, where we find the similarity between the keyphrases by first extracting the keyphrase embeddings, then computing similarity with existing keyphrases in the keyphrase store, and finally deciding whether the keyphrase should be stored. However, we observe that using fuzzy search in the pipeline produces more diverse keyphrases compared to PhraseBERT.
\\\\
Our keyphrase sampling pipeline is described in Figure~\ref{fig:keyphrase-sampling-pipeline} and Figure~\ref{fig:keyphrase-sampling-example} in Appendix.
\subsection{Entity Sampler}
To address the problem of text normalization, we create an entity generator that not only generates the semiotic classes but also their normalized forms. Our entity sampler can generate complex, real-world variations and combinations of semiotic classes. Since the rules for generating the entities are encoded in the entity sampler, normalization occurs simultaneously with generation. This approach ensures deterministic generation with guaranteed accuracy in normalization, as the entities do not yet exist in the text. For example, we might generate an email address composed of a first name, a last name separated by an underscore, and random characters. This allows us to normalize the email address while these components are being concatenated. Our entity sampler is capable of generating thousands of unique combinations across 9 different entities: \emph{Addresses, Phone Numbers, Email Addresses, URLs, Dates, Times, Percentages, Person Names with Salutations}. Our entity sampler is also locale-sensitive and multilingual. In Appendix, Figure~\ref{fig:entity-recipies} describes the recipes for entity generation and normalization for different semiotic classes, while Table~\ref{tab:entity-examples-english} and Table~\ref{tab:entity-examples-spanish} contain different examples of such classes with their normalized forms.

\subsection{Text Script Generator}
We combine the generated keyphrases with the semiotic classes in a prompt to generate domain-specific text. We use lm-format-enforcer to force the model to generate the text in JSON format, ensuring that only the required text scripts are generated. We also replace the semiotic classes in the text with their normalized forms to generate the normalized script. Using different prompts, we can generate various sentence types for our text scripts. Table~\ref{tab:generation-prompts} in Appendix shows different prompts used for generating text scripts for different sentence types.

\subsection{Normalization Post Processing}
LLM-generated text may occasionally introduce new semiotic classes. Therefore, we use a basic post-processing algorithm to normalize the text. The algorithm expands the acronyms, converts numbers to their cardinal forms, and removes any hyphens, underscores, and brackets from the normalized script. Our analysis reveals that post-processing steps, such as changing numbers non-contextually to their cardinal forms, may introduce normalization errors. However, given that the scripts we generate are small (upto 50 words) the occurrence of such errors is quite rare, and our overall process still achieves high normalization accuracy (Section~\ref{sec:normalization-accuracy}).
\subsection{Speech Audio Generation And Cross Lingual Voice Cloning}
Once the text and its normalized forms are generated, we feed the normalized text to the Speech Audio Generation Module. The audio generation module takes in the input text, and a reference audio, and generates speech audio with voice cloned as per the reference audio. We first generate a base speech audio using a pretrained TTS model \cite{zhao2024melo}. Then,
for speaker standardization, we use OpenVoiceV2's \cite{qin2024openvoiceversatileinstantvoice} tone color converter with reference voices taken from proprietary voice artists. The tone color converter is language agnostic i.e. we can use reference audio in English to standardize voices in other languages. This allows us to use standard voice artists across languages for our downstream TTS system.

Data samples generated using SpeechWeave are available in Table~\ref{tab:generated-samples}.
\section{Evaluation}
\label{sec:evaluation}
To evaluate our pipeline, we generate a dataset with 3000 datapoints across 16 business domains, 5 sentence types, 9 semiotic classes, and 2 reference speakers (male and female), each in English and Spanish. Sentences with fewer than 5 or more than 50 words are excluded and regenerated using a different seed. For the baseline (wherever applicable), we prompt a large language model to generate text in the required business domain, as detailed in Table~\ref{tab:generation-prompts} in Appendix. For diversity evaluation, we also compare our results to public datasets — English Librispeech \cite{7178964} and Spanish LibriSpeech \cite{Pratap_2020}  - sampling 3000 datapoints from each, applying the same filtering criteria. For evaluating the quality of a downstream model trained on our dataset, we use the test splits from the same public datasets. Experiment settings are detailed in Appendix~\nameref{sec:appendix-experimentation-settings}.

\subsection{Diversity Analysis}
\label{sec:diversity-analysis}
For diversity analysis, we examine the variation in both the generated text and speech across different samples produced by the pipeline.
\subsubsection{Diphone Coverage}
Diphones are adjacent phonemes representing transitions in speech, and diphone coverage indicates how well a corpus captures phoneme combinations. Our results show that relatively, our pipeline's data covers \textbf{17.4\%} more diphones in English and \textbf{9.7\%} more in Spanish compared to the baseline. However, the public LibriSpeech datasets cover \textbf{5.7\%} more in English and \textbf{15.2\%} more in Spanish than our pipeline's data. The superior coverage in LibriSpeech can largely be attributed to high mean word count compared to our dataset. Experimentation settings and diphone coverage comparisons are provided in Appendix~\ref{sec:appendix-diphone-coverage} and Figure~\ref{fig:diphone-coverage} respectively.

\subsubsection{Mean Pairwise Similarity}
We evaluated the semantic mean pairwise similarity within sentence groups categorized by business domain and type. Compared to direct prompting, our pipeline generates more diverse text, showing relatively \textbf{45.8\%} and \textbf{44.4\%} lower grouped similarity scores for English and Spanish, respectively. Even in the most homogeneous group, our data's similarity scores were relatively \textbf{48.5\%} and \textbf{46.7\%} lower for English and Spanish compared to the baseline.
Since public speech datasets aren't categorized by business domain, we calculated mean pairwise similarity without grouping for comparison. Our dataset shows greater diversity, with relative mean similarity scores lower by \textbf{58.8\%} for English and \textbf{10.7\%} for Spanish.
Experimentation settings and some additional analysis are described in Appendix~\ref{sec:appendix-evaluation-pariwise-sim}
\subsubsection{Token Diversity}
Token diversity, measured by Type-Token Ratio (TTR) and Moving Average Type-Token Ratio (MATTR), reflects lexical richness. Our results show both TTR and MATTR are higher in our synthesized dataset compared to LLM-generated text and both public datasets - LibriSpeech English and LibriSpeech Spanish.

Table~\ref{tab:merged-comparison} contains a comparison of the datasets on these diversity indicators.

\subsection{Quality Analysis}
The quality of the data generated by our pipeline is assessed across three key dimensions: Normalization Accuracy, Speech Audio Clarity and Downstream Model Training.

\subsubsection{Normalization Accuracy}
\label{sec:normalization-accuracy}
We evaluate our at-source normalization technique against \emph{Nvidia NeMo's text normalizer} \cite{zhang2021nemoinversetextnormalization}. Normalization accuracy is the ratio of correctly normalized sentences to the total evaluated. Our pipeline achieves \textbf{0.97} and \textbf{0.94} for English and Spanish, while NeMo scores \textbf{0.67} and \textbf{0.54}, showing superior performance of at-source text normalization for training data generation. NeMo’s errors involve mishandling variations of semiotic classes, such as breaking up names, improperly normalizing phone numbers, or missing alternate currencies. Experimentation settings are detailed in Appendix Section~\ref{sec:appendix-normalization-accuracy}.
\begin{table}
  \centering
  \small
  \renewcommand{\arraystretch}{1.3}
\begin{tabular}{p{1.5cm} p{2cm} p{2.5cm}}
    \hline
    \textbf{Language} & \textbf{Technique} & \textbf{\begin{tabular}[c]{@{}l@{}}Normalization \\ Accuracy\end{tabular}} \\
    \hline
    \multirow{2}{*}{English} & NeMo & 0.67 \\
    & Ours & \textbf{0.97} \\
    \hline
    \multirow{2}{*}{Spanish} & NeMo & 0.54 \\
    & Ours & \textbf{0.94} \\
    \hline
  \end{tabular}
  \caption{Comparison of our at-source text normalization accuracy with Nemo’s Text Normalizer.}
  \label{tab:normalization-accuracy}
\end{table}

\subsubsection{Speech Audio Clarity} 
We evaluate the acoustic quality of the generated speech to assess the effectiveness of the pretrained model in synthesizing speech from our pipeline's text scripts. Performance is quantified using Mean Signal-to-Noise Ratio (SNR), Automated Mean Opinion Score (MOS), and Word Error Rate (WER).  

\begin{table}[htbp]
  \centering
  \small
  \renewcommand{\arraystretch}{1.3}
\begin{tabular}{p{1.5cm}p{1.2cm} p{1.2cm}p{1.2cm}}
    \hline
    \textbf{Language} & \textbf{SNR (dB)} & \textbf{MOS} & \textbf{WER (\%)} \\
    \hline
    \multirow{1}{*}{English} & 59.82 & 4.95 & 9.32 \\
    \hline
    \multirow{1}{*}{Spanish} & 53.01 & 4.87 & 15.21 \\
    \hline
  \end{tabular}
  \caption{Speech audio clarity indicators for the data generated by SpeechWeave}
  \label{tab:snr-mos-wer}
\end{table}

Table~\ref{tab:snr-mos-wer} shows that the synthesized speech achieves high MOS and SNR scores with low WER, demonstrating superior audio quality and strong textual and phonetic accuracy. Experimentation settings available in Appendix~\ref{sec:appendix-speech-audio-clarity}.

\subsubsection{Downstream Model Training}
We fine-tuned a \emph{StyleTTS 2} model \cite{li2023styletts2humanleveltexttospeech} using a \emph{LibriTTS-trained checkpoint} on data generated by our pipeline and evaluated its quality using \emph{WER}. As a baseline, we measured WER on the LibriSpeech test dataset \cite{7178964, Pratap_2020} before fine-tuning. Our results show significant WER reductions: \textbf{40\%} for English and \textbf{27\%} for Spanish relatively, compared to the baseline, demonstrating the effectiveness of our pipeline in generating high-quality training data for improved speech synthesis.  

\begin{table}[htbp]
  \centering
  \small
  \renewcommand{\arraystretch}{1.3}
  \begin{tabular}{p{3cm}p{1.7cm} p{1.7cm}}
    \hline
    \textbf{\centering Model} & \textbf{\centering LibriSpeech English WER (\%)} & \textbf{\centering LibriSpeech Spanish WER (\%)} \\
    \hline
    LibriTTS Checkpoint (Baseline) & 15.37 & 85.05 \\
    \hline
    Baseline fine-tuned on our data & \textbf{9.36} & \textbf{48.44} \\
    \hline
  \end{tabular}
  \caption{WER before and after fine-tuning StyleTTS 2 with SpeechWeave-generated data}
  \label{tab:downstream-training}
\end{table}

Table~\ref{tab:downstream-training} summarizes the experimental results with experimentation settings described in Appendix Section~\ref{sec:appendix-downstream-model-training}. It's worth noting that StyleTTS 2 does not have a Spanish-trained checkpoint, which explains the higher overall WER for Spanish. In this context, training on our Spanish data effectively adapts the model to the Spanish language.

\section{Conclusion}
We introduce \textbf{SpeechWeave}, a simple yet effective pipeline for generating diverse, normalized text and speaker standarized speech audio data for training text to speech systems. Our analysis reveals that the data generated by our pipeline is much more diverse than the data generated by directly prompting an LLM, and carries higher normalization accuracy compared to post processing normalizers like NeMo while being speaker-standardized to allow scaling training data. The data is also on par with publicly available speech datasets, while adhering to the required business domains. Given that the data is highly precise in terms of normalization, it can also be used to train text normalization models.
\section*{Limitations and Future Work}
The accuracy of the normalized text generated by our pipeline is limited by the number of semiotic classes supported by the the entity sampler. Moreover, although our pipeline incorporates \emph{Mistral-7b-Instruct-0.3} and \emph{OpenVoice V2} Stack for data generation, the results may vary depending upon the models chosen for generating the dataset. Our evaluation is also limited to English and Spanish languages and the extent of improvement may vary based on the language for which the data is generated. In the future, we plan to extend the framework to include other morphologically rich languages, with a particular focus on those that are currently underrepresented. Moreover, while, it is fairly straightforward to support a new semiotic class, the post processor may result in occasional normalization errors for unsupported entities. We wish to continue this work by generalizing the framework for semiotic class generation and entity normalization at source. We would also like to extend this work to support styled speech audio generation and speech style standardization. 

\section*{Ethical Considerations}
Our work uses entirely synthetic text and audio data generated through a controlled pipeline, without the involvement of real-world user data or human participants, apart from publicly available speech datasets used solely for evaluation purposes. This design inherently avoids privacy violations and ensures that no personally identifiable information is processed or exposed. As such, our data generation process does not pose significant ethical risks typically associated with data collection, consent, or user harm. By relying on synthetic data, we uphold best practices in privacy-preserving and ethically responsible research.
\section*{Acknowledgments}
The work was conducted during employment with and funded by Oracle Corporation (AI Services).

\bibliography{custom}
\section*{Appendices}
\appendix
\section{Generated Samples}
Table~\ref{tab:generated-samples} contains examples of text scripts generated by our pipeline along with their normalized forms.

\section{Keyphrase Sampling Pipeline}
\label{sec:appendix-keyphrase-sampling}
To generate keyphrases to increase diversity in the generated text scripts, we prompt the LLM in multiple steps. We begin by prompting the LLM to generate subdomains in the required business domain. Then we prompt the LLM to pick one subdomain randomly. Then the LLM is required to write a creative paragraph about the subdomain in the target language. Finally, we prompt the LLM to extract keyphrases from the generated paragraph. Output formats are enforced using lm-format-enforcer \cite{Gat} and set at different steps in the prompt chain.
For each of the generated keyphrases, we determine the similarity score with the rest of the keyphrases generated by the pipeline (grouped by domain and language) using Token Sort Ratio. Any keyphrase that has a token sort ratio of less than 0.8 is then stored in the keyphrase store. The process is repeated unless the required number of keyphrases is available in the store. For the conducted evaluation experiments, we use two keyphrases per text script. Figure~\ref{fig:keyphrase-sampling-pipeline} describes the keyphrase sampling pipeline, while Figure~\ref{fig:keyphrase-sampling-example} depicts an example at each step of the prompt chain.

\section{Entity Sampling}
Our entity sampler consists of recipes to generate several forms of semiotic classes along with their normalized forms. The sampler consists of recipes for each language and is extensible to support more languages. In most of the scenarios, the base entities are generated using the Faker library \cite{joke2kfaker}. For example, for generating an email, person names are generated using Faker library \cite{panda2025techniques, agarwal-etal-2025-fs, agarwal2024domain}. The exact recipes for different entity and their forms are described in Figure~\ref{fig:entity-recipies} and some example of generated entities and their normalized forms are present in Table~\ref{tab:entity-examples-english} and Table~\ref{tab:entity-examples-spanish}.

\section{Text Script Generation Pipeline}
\label{sec:appendix-text-script}
Entire text script generation pipeline is described in Figure~\ref{fig:text-script-pipeline}.
\section{Experimentation Settings}
\label{sec:appendix-experimentation-settings}
\subsection{Keyphrases and Text Scripts Generation}
The keyphrases and text scripts are generated using Mistral-7b-Instruct-0.3 \cite{jiang2023mistral7b} model with a temperature setting of 1.2, a top\_p value of 0.9. The data generated by the baseline technique shares characteristics with the data produced by our pipeline, including the use of the same LLM, dataset size, business domains, sentence types, sampling parameters, and length filtering criteria.

\subsection{Evaluation}
\subsubsection{Diphone Coverage}
\label{sec:appendix-diphone-coverage}
To estimate the diphone coverage in our dataset and compare it with baseline corpora, we begin by extracting all unique phonemes from the text scripts using a phonemizer \cite{g2pE2019, patel2025sweeval, Bernard2021}. After identifying the phonemes, we compute the diphones by examining each pair of adjacent phonemes. Figure~\ref{fig:diphone-coverage} depicts the diphone coverage for different dataset sizes for the three datasets we compared.
\subsubsection{Pairwise Similarity}
\label{sec:appendix-evaluation-pariwise-sim}
Since the generated text data is domain-specific, we compute mean pairwise similarity \cite{Gong_2019, thomas2025model} within sentence groups categorized by business domain and sentence type. Specifically, the dataset is first segmented based on these categories, and the mean pairwise similarity is then calculated within each group. A global similarity score (\ref{global-group-sim}) is obtained by averaging these group-level similarity scores. The embeddings for calculating this metric are obtained using the LaBSE model \cite{feng2022languageagnosticbertsentenceembedding}.
\begin{equation}
\label{global-group-sim}
\tiny
\text{Grouped Similarity} = \frac{1}{|G|} \sum_{g \in G} \left( \frac{1}{|S_g|(|S_g|-1)} \sum_{i=1}^{|S_g|} \sum_{j=i+1}^{|S_g|} \cos(s_i^g, s_j^g) \right)
\end{equation}
where $|G|$ is the total number of groups, $|S_g|$ is the number of sentences in group $g$, and $s_i^g$ and $s_j^g$ are LaBSE embeddings for sentences at indices $i$ and $j$ in group $g$.

Objectively, our pipeline produces significantly better results than the direct prompting baseline. A quick manual review also reveals that the direct prompting pipeline tends to generate sentences excessively centered around certain phrases. For example: (1) 23\% of sentences generated in the Banking domain contain the phrase "savings account," compared to just 1.8\% in our pipeline. (2) 14\% of all sentences generated in the Finance domain contain the phrase "stock market," compared to just 0.5\% in our pipeline.

Non Grouped mean pairwise similarity is calculated as per Equation~\ref{non-group-pairwise-sim}.
\begin{equation}
\label{non-group-pairwise-sim}
\small
\text{Non Group Similarity} = \frac{1}{M(M-1)} \sum_{j=1}^{M} \sum_{k=j+1}^{M} \cos(s_j, s_k)
\end{equation}

where \( M \) is the total number of sentences, \( s_j \) and \( s_k \) are embeddings for sentences at index \( j \) and \( k \) in the group.

\subsubsection{Token Diversity}
\label{sec:appendix-token-diversity}
To compute TTR, MATTR, we first tokenize the text using NLTK's \emph{Punkt} tokenizer \cite{Bird_Natural_Language_Processing_2009} and SpaCy's \emph{es\_core\_news\_sm} model \cite{spacy2, pattnayak-etal-2025-hybrid} for Spanish text processing.

\begin{itemize}
    \item TTR (Type-Token Ratio): Calculated as the ratio of the number of unique tokens to the total number of tokens in the text.  
    \item MATTR (Moving Average Type-Token Ratio): Calculated as TTR over a sliding window of size 100, and then averaging the values.  
\end{itemize}
\subsubsection{Normalization Accuracy}
\label{sec:appendix-normalization-accuracy}
While our pipeline performs at-source text normalization along with some basic post-processing steps, we observe that certain semiotic classes generated by the large language model (which we didn't use in our prompt) may not be correctly normalized. These normalization errors stem from either the absence or incorrect application of normalization to these new semiotic classes. To establish a ground truth for assessing normalization accuracy, we manually evaluate 500 (each for English and Spanish) sentences generated by our pipeline. For any incorrectly normalized sentence, the correct normalization is documented and used as the ground truth.

To further assess the performance of our technique, we apply Nvidia NeMo's WFST text normalizer to the generated sentences. We note that NeMo's text normalizer fails to perform certain fundamental normalization tasks, such as removing hyphens or expanding acronyms, which are handled by our pipeline’s postprocessor. To mitigate errors arising from these discrepancies, we apply the same postprocessor to NeMo's output. Additionally, we observe that NeMo follows a different strategy for normalizing phone numbers, specifically regarding the placement of commas, compared to our pipeline. As such, we exclude comma placement from penalties. A sentence is considered penalized if its output does not match the ground truth. We also recognize that NeMo may produce outputs that differ from our normalization process but are still acceptable. To avoid penalizing these differences, we manually review all penalized instances and classify those with acceptable normalization as correct. A couple of examples of such acceptable errors are: (1) Incorrect deduction of locale for normalizing dates. For example, normalizing 02-01-2005 as "February one, twenty twenty five" instead of "January two twenty twenty five" as done by our pipeline. (2) Normalizing large amounts with "and" separator. For example, normalizing \$301,000 as "three hundred one thousand" instead of "three hundred and one thousand" as done by our pipeline.
\subsubsection{Speech Audio Clarity}
\label{sec:appendix-speech-audio-clarity}
\begin{itemize}
    \item Mean Opinion Score (MOS): We estimated MOS using the NISQA (Neural Speech Quality Assessment) model \cite{mittag21_interspeech}, which predicts speech quality based on perceptual metrics without requiring human evaluation.
    
    \item Signal-to-Noise Ratio (SNR): measures the level of speech signal relative to background noise. It is calculated as:  
    \begin{equation}
        SNR = 10 \log_{10} \left( \frac{P_{\text{signal}}}{P_{\text{noise}}} \right)
    \end{equation}
    where \( P_{\text{signal}} \) represents the power of the speech signal, and \( P_{\text{noise}} \) represents the power of background noise. A higher SNR indicates cleaner audio with less noise interference. Since we lack a reference clean audio, we estimated the noise power from the quietest segments of the audio, assuming that these portions (where no speaker is present) primarily contain background noise.
    
    \item Word Error Rate (WER):  We utilized WER as a metric to measure how accurately the synthesized audio samples reflect the original normalized text, effectively evaluating the performance of the pipeline generating audio from the normalized text . This is achieved by leveraging an ASR model \cite{nvidia_riva_tts, nemo2021} to transcribe the synthesized audio samples. We then compute the WER by comparing the transcribed text to the source normalized text. \newline
It is calculated as:  
    \begin{equation}
        WER = \frac{S + D + I}{N} \times 100
    \end{equation}
    where:
    \begin{itemize}
        \item \( S \) is the number of substitutions (incorrect words),
        \item \( D \) is the number of deletions (missing words),
        \item \( I \) is the number of insertions (extra words), and
        \item \( N \) is the total number of words in the reference text.
    \end{itemize}

A lower WER indicates that the synthesized audio samples accurately reflect the input normalized source text.
\end{itemize}
\subsubsection{Downstream Model Training}
\label{sec:appendix-downstream-model-training}
To evaluate the effectiveness of the synthetic dataset generated by our pipeline for real-world Text-to-Speech synthesis, we conducted downstream model training using the StyleTTS 2 model. We began by using a StyleTTS' LibriTTS checkpoint as our base model.

For the baseline setup, we performed inference on the LibriSpeech test datasets, which are out-of-distribution (OOD) with respect to both our generated dataset and the LibriTTS training data. Test set contains 2618 samples for English and 2385 samples for Spanish. These text inputs were passed through the baseline model to synthesize speech audios.

We then evaluated the synthesized audio using NVIDIA NeMo's automatic speech recognition (ASR) models: stt\_en\_conformer\_ctc\_large for English and stt\_es\_conformer\_ctc\_large for Spanish \cite{nvidia_riva_tts}. These ASR models transcribed the generated audio into text, which was then compared to the reference input using {Word Error Rate (WER) as the evaluation metric. To ensure fair and robust evaluation, we used a reference speaker audio that was not present in the training set for both the baseline and fine-tuned models.

For fine-tuning, we trained StyleTTS 2 models using the pipeline-generated datasets for English and Spanish, initializing from the same LibriTTS checkpoint and training for 50 epochs. The training uses PLBERT \cite{li2023phonemelevelbertenhancedprosody} for English and  a multilingual variant of the same for Spanish for grapheme predictions.

\section{Text Script Generation Prompts}
\label{sec:generation-prompts}
Prompts use for generating text scripts using direct prompting and through our pipeline are available in Table~\ref{tab:generation-prompts}

\section{A note on secondary seeds}
\begin{itemize}
    \item Reproducibility is an essential component in any machine learning pipeline. For text generation, we need to ensure that the generated dataset is reproducible.
    \item We have stochastic components in our pipeline, such as Random Entity Generator, which can cause the entire pipeline to generate different text if not controlled.
    \item Large Language Models also have stochastic components that cause them to generate different text even when the inputs remain the same.
    \item One common way to control the stochasticity of both these components is by fixing the random seed. This ensures a component follows the same path when run again and again.
    \item However, fixing this seed is a limitation for us. There may be situations where we need to generate something in a loop. For example:
    \begin{itemize}
        \item We may need to generate 5 email addresses. If we fix the seed, we will get the same value repeatedly.
        \item When filtering a sentence based on some criteria (e.g., it is too long), generating the sentence using the same seed will keep producing the same sentence.
    \end{itemize}
    \item To eliminate this, we use a process called secondary seeding.
    \item We first generate a primary seed and fix it. With the primary seed fixed, we generate a secondary seed anytime we need to run a random generation.
    \begin{itemize}
        \item For example, if we encounter a generated sentence that is too long and needs to be filtered, we generate a new secondary seed. This generates a new sentence different from the last one.
    \end{itemize}
    \item Secondary Seeding also ensures reproducibility. Since the secondary seed is generated using the primary seed, the sequence of secondary seed generation remains the same.
    \item Therefore, if you run the pipeline using the same primary seed again, you will generate the same data.
\end{itemize}
Secondary seeding is described in Figure~\ref{fig:secondary-seed}

\begin{table*}[t]
    \small
  \renewcommand{\arraystretch}{1.3} 
  \begin{tabular}{p{7.5cm}p{7.5cm}}
    \hline
    \textbf{Text Script} & \textbf{Normalized Form} \\
    \hline
    Mrs. Julie Young was blown away by the sheer size of the aircraft and the luxurious amenities offered by the airline! & Missis Julie Young was blown away by the sheer size of the aircraft and the luxurious amenities offered by the airline! \\
    \hline
    I'll be reaching out to Abigail Walker at 5.abigail.walker@yandex.com to discuss this further. & I'll be reaching out to Abigail Walker at five dot abigail dot walker at yandex dot com to discuss this further. \\
    \hline
    With 87\% of repair manuals available online in step-by-step instructions, maintenance and repairs on automobiles have become more accessible and efficient. & With eighty seven percent of repair manuals available online in step by step instructions, maintenance and repairs on automobiles have become more accessible and efficient. \\
    \hline
    Dr. Angel Roberts has made it easier for customers to make major purchases by simplifying the process and reducing the necessary steps. & Doctor Angel Roberts has made it easier for customers to make major purchases by simplifying the process and reducing the necessary steps. \\
    \hline
    The city council is working on delivering a new £273 million scheme to improve the built environment for its residents. & The city council is working on delivering a new two hundred and seventy three million pounds scheme to improve the built environment for its residents. \\
    \hline
    El 02-01-1997 fue la fecha en la que Desmarca abrió su tienda, con un fuerte énfasis en la personalización de los productos. & El dos de enero de mil novecientos noventa y siete fue la fecha en la que Desmarca abrió su tienda, con un fuerte énfasis en la personalización de los productos. \\
    \hline
    El sistema de control de vuelo utiliza una señal de posición con un 93,45\% de precisión para determinar la ubicación de la aeronave sobre la Tierra. & El sistema de control de vuelo utiliza una señal de posición con un noventa y tres coma cuarenta y cinco por ciento de precisión para determinar la ubicación de la aeronave sobre la Tierra. \\
    \hline
    ¿Has realizado un análisis financiero de los instrumentos financieros que están disponibles para invertir CA\$572? & ¿Has realizado un análisis financiero de los instrumentos financieros que están disponibles para invertir quinientos setenta y dos dólares canadienses? \\
    \hline
    El informe sobre la corrupción en el gobierno se puede consultar en 86corrupti.net. & El informe sobre la corrupción en el gobierno se puede consultar en ocho seis corrupti punto net. \\
    \hline
  \end{tabular}
  \caption{Examples of generated text scripts and their normalized forms.}
  \label{tab:generated-samples}
\end{table*}
\begin{figure*}
  \includegraphics[width=1.0\linewidth]{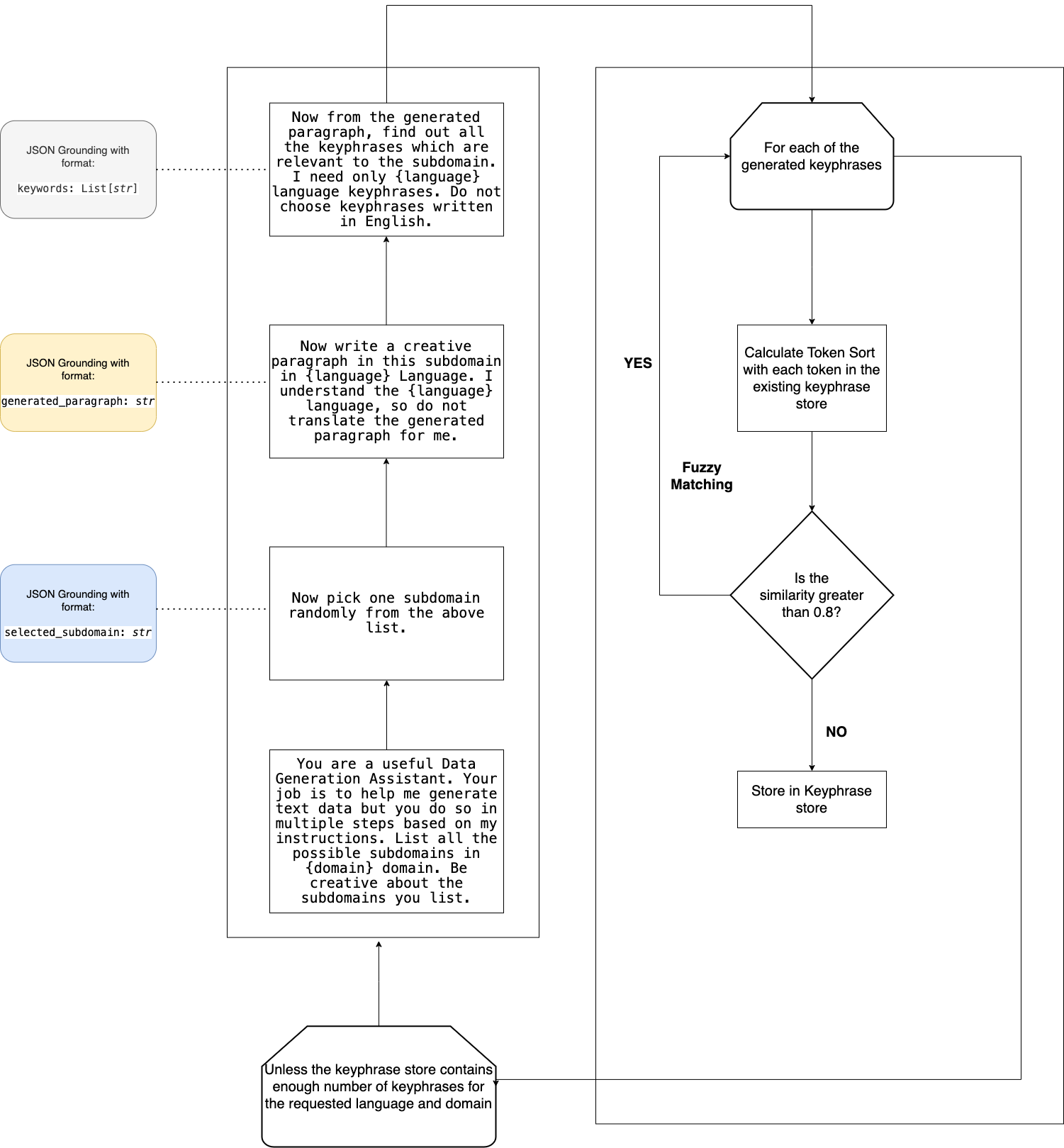}
  \caption {\label{fig:keyphrase-sampling-pipeline}Multistep Keyphrase Sampling Pipeline with De-duplication and Keyphrase Store}
\end{figure*}
\begin{figure*}
  \includegraphics[width=1.0\linewidth]{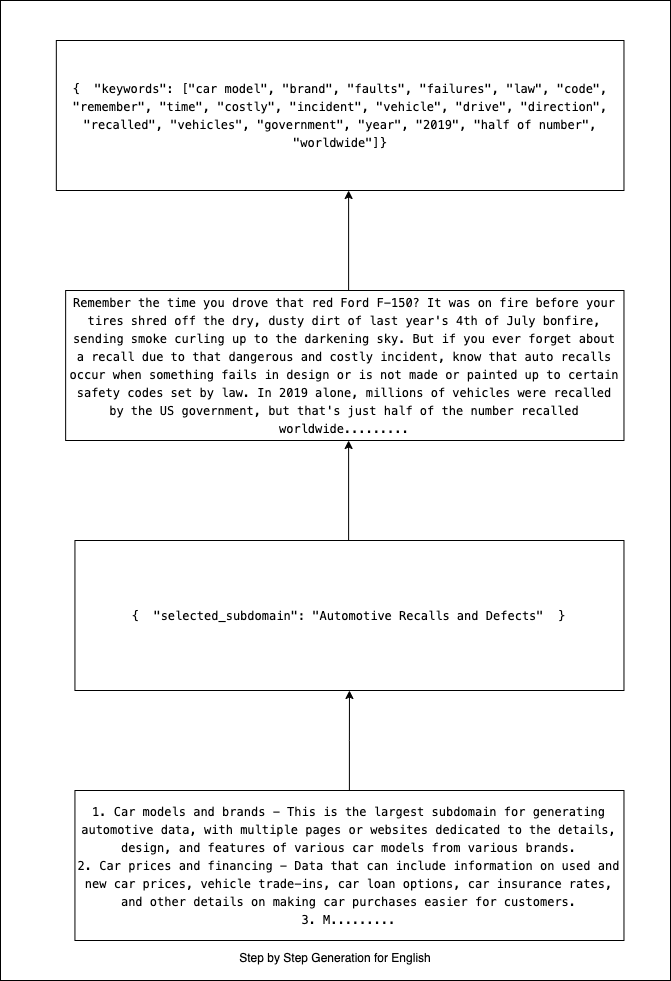}
  \caption {\label{fig:keyphrase-sampling-example}Example output from keyphrase sampling pipeline at each step of the prompt chain}
\end{figure*}

\begin{table*}[t]
  \centering
    \small
  \renewcommand{\arraystretch}{1.3} 
  \begin{tabular}{p{2cm}p{5.2cm}p{7cm}}
    \hline
    \textbf{Type} & \textbf{Generated Entity} & \textbf{Normalized Form} \\
    \hline
    Amount & 863k Canadian Dollars & Eight Hundred and Sixty Three Thousand Canadian Dollars \\
           & 29 USD & Twenty Nine U S Dollars \\
           & £723m & Seven Hundred and Twenty Three Million Pounds \\
    \hline
    Date   & 10-04-2023  & October fourth twenty twent three \\
           & 10/21/1997  & October twenty first ninet seven \\
           & 06/Jan/10   & January sixth ten \\
    \hline
    Person & Dr. Yvette Nelson  & Doctor Yvette Nelson \\
           & Mr. Cameron Carter  & Mister Cameron Carter \\
           & Mrs. Julia Thomas   & Missis Julia Thomas \\
    \hline
    Email  & cbrwthomaswalker29@hotmail.com & c b r w thomas walker two nine at hot mail dot com \\
           & l51sonyasanders@mail.com & l five one sonya sanders at mail dot com \\
    \hline
    Phone Number & 7854017402 & seven eight five, four zero one, seven four zero two \\
                 & +1-47859964121 & plus one, four seven eight five, nine nine six, four one two one \\
    \hline
    Percentage & 39.29\% & thirty nine point two nine percent \\
    \hline
    URL & \emph{http://though15.eu} & h t t p colon slash slash though one five dot e u \\
    \hline
    Address & Johnson Trail Plz KY 45287 & Johnson Trail Plaza Kentucky four five two eight seven \\
            & Chen Inlet North Dakota 34101 & Chen Inlet North Dakota three four one zero one \\
    \hline
    Time & 13:59 & Thirteen fifty nine \\
         & 17:00 & Seventeen hundred hours \\
         & 02:34 PM & Two thirty four P M \\
         & 11 o'clock & Eleven o clock \\
    \hline
  \end{tabular}
  \caption{\label{tab:entity-examples-english}Examples of generated entities and their normalized forms across various semiotic classes in English.}
  \label{tab:normalized_entities}
\end{table*}
\begin{table*}
  \centering
  \small
  \renewcommand{\arraystretch}{1.3} 
  \begin{tabular}{p{2cm}p{5.2cm}p{7cm}}
    \hline
    \textbf{Type} & \textbf{Generated Entity} & \textbf{Normalized Form} \\
    \hline
    Amount & CA\$572 & quinientos setenta y dos dólares canadienses \\
           & A\$485,986,561.71 & cuatrocientos ochenta y cinco millones novecientos ochenta y seis mil quinientos sesenta y uno con setenta y un centavos dólares australianos \\
           & £723m & setecientos veintitrés millones de libras \\
    \hline
    Date   & 05/22/93 & veintidós de mayo de mil novecientos noventa y tres \\
           & 02-Oct-1988 & dos de octubre de mil novecientos ochenta y ocho \\
           & 08-04-2000 & ocho de abril de dos mil \\
    \hline
    Person & Prof. Edgardo Aragón Trujillo & El Profesor Edgardo Aragón Trujillo \\
           & Dr. Bernabé Quintanilla Cerezo & El Doctor Bernabé Quintanilla Cerezo \\
           & Sr. Rodolfo del Cid & El Señor Rodolfo del Cid \\
    \hline
    Email  & 16rosaliaquesada@outlook.com & uno seis rosalia quesada en outlook punto com \\
           & ferreraclara36@outlook.com & ferrera clara tres seis en outlook punto com \\
    \hline
    Phone Number & 4 835600765 & cuatro ocho tres, cinco seis cero, cero siete seis cinco \\
                 & 4807 14 77 34 & cuatro ocho cero, siete uno cuatro, siete siete tres cuatro \\
    \hline
    Percentage & 69.76\% & sesenta y nueve punto setenta y seis por ciento \\
               & 76\% & setenta y seis por ciento \\
    \hline
    URL & 73corporis.gov & siete tres corporis punto gov \\
    \hline
    Address & 79 Pasaje de Claudio Jimenez Vlg Tarragona Colorado 11282 & siete nueve Pasaje de Claudio Jiménez Aldea Tarragona Colorado uno uno dos ocho dos \\
            & Pasadizo Julián Bosch Louisiana 32198 & Pasadizo Julián Bosch Louisiana tres dos uno nueve ocho \\
    \hline
    Time & 09:20 & nueve veinte \\
         & 07:59 pm & siete cincuenta y nueve p m \\
         & las 2 en punto & las dos en punto \\
    \hline
  \end{tabular}
  \caption{\label{tab:entity-examples-spanish}Examples of generated entities and their normalized forms across various semiotic classes in Spanish.}
  \label{tab:normalized_entities_spanish}
\end{table*}

\begin{table*}
  \centering
  \small
  \renewcommand{\arraystretch}{1.3}
  \begin{tabular}{p{3.5cm} p{2cm} p{6cm}}
    \hline
    \textbf{Dataset} & \textbf{Sentence Type} & \textbf{Prompt} \\
    \hline
    \multirow{5}{*}{Direct Prompting (Baseline)} 
    & Statement & Construct one sentence in \{language\} language in \{domain\} domain. I am well aware of \{language\} language, so do not translate it. \\ 
    & Exclamation & Construct one sentence in \{language\} language in \{domain\} domain. The generated sentence should be exclamatory and have a surprising tone. I am well aware of \{language\} language, so do not translate it. \\ 
    & Question & Construct one sentence in \{language\} language in \{domain\} domain. The generated sentence should be a question. I am well aware of \{language\} language, so do not translate it. \\ 
    & Phrase & Construct a short phrase in \{language\} language in \{domain\} domain. The phrase should contain about 5 to 7 words. It should be strictly a phrase and not a sentence. I am well aware of \{language\} language, so do not translate it. \\ 
    & Utterance & Construct a short arbitrary conversation between two people in \{language\} language in \{domain\} domain. I am well aware of \{language\} language, so do not translate it. \\ 
    \hline
    \multirow{5}{*}{Ours} 
    & Statement & Construct one sentence in \{language\} language in \{domain\} domain with the following words: \{words\}. The following entities should also be present in the text: \{entities\}. \\ 
    & Exclamation & Construct one sentence in \{language\} language in \{domain\} domain with the following words: \{words\}. The following entities should also be present in the text: \{entities\}. The generated sentence should be exclamatory and have a surprising tone. \\ 
    & Question & Construct one sentence in \{language\} language in \{domain\} domain with the following words: \{words\}. The following entities should also be present in the text: \{entities\}. The generated sentence should be a question. \\ 
    & Phrase & Construct a short phrase in \{language\} language in \{domain\} domain with the following words: \{words\}. The phrase should contain about 5 to 7 words. The phrase should not have any numbers or dates. It should be strictly a phrase and not a sentence. \\ 
    & Utterance & Construct a short arbitrary conversation between two people in \{language\} language in \{domain\} domain containing the following words: \{words\}. The following entities should also be present in the text: \{entities\}. \\ 
    \hline
  \end{tabular}
  \caption{\label{tab:generation-prompts}
    Prompts used for Text Generation through direct prompting (baseline) and our pipeline.
  }
\end{table*}

\begin{figure*}
  \includegraphics[width=1.0\linewidth]{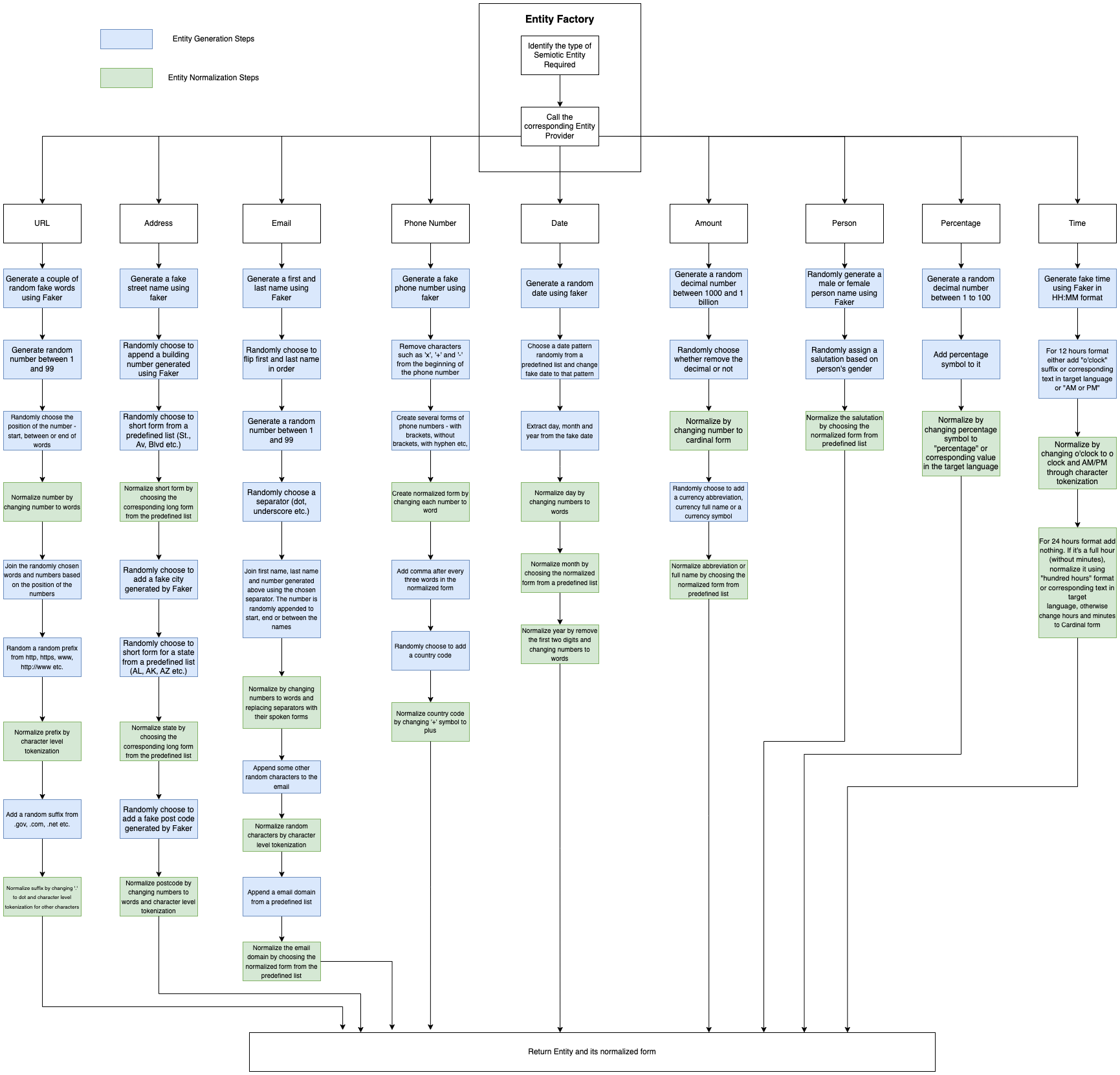}
  \caption {\label{fig:entity-recipies}Recipes for generating different semiotic classes and their normalized forms}
\end{figure*}
\begin{figure*}[htbp]
  \includegraphics[width=1.0\linewidth]{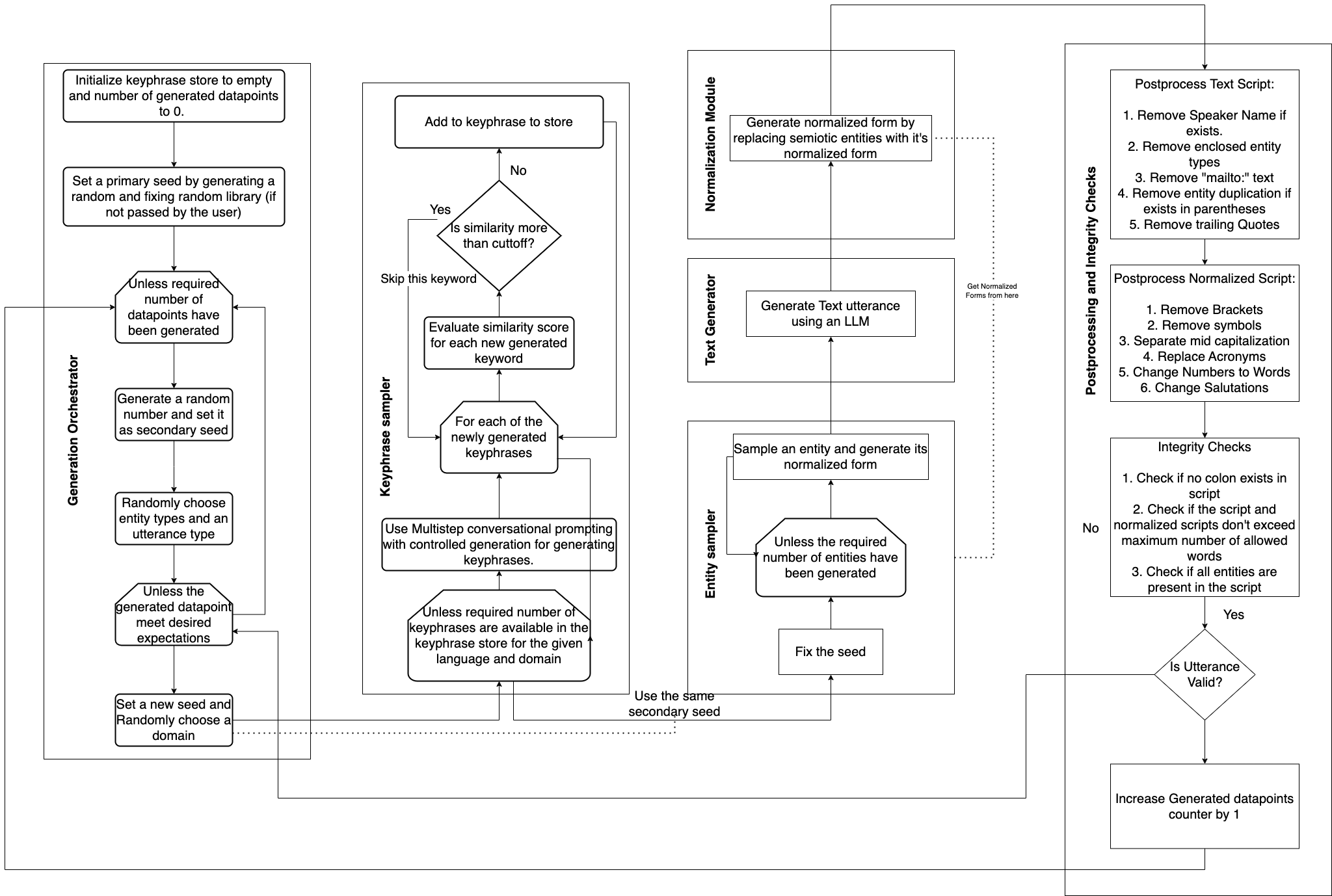}
  \caption {\label{fig:text-script-pipeline}Detailed description of text script generation pipeline}
\end{figure*}
\begin{figure*}[t]
  \includegraphics[width=1\linewidth]{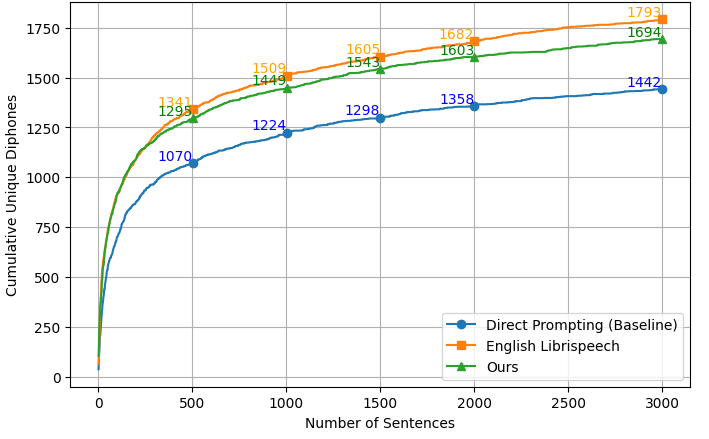} \hfill
  \includegraphics[width=1\linewidth]{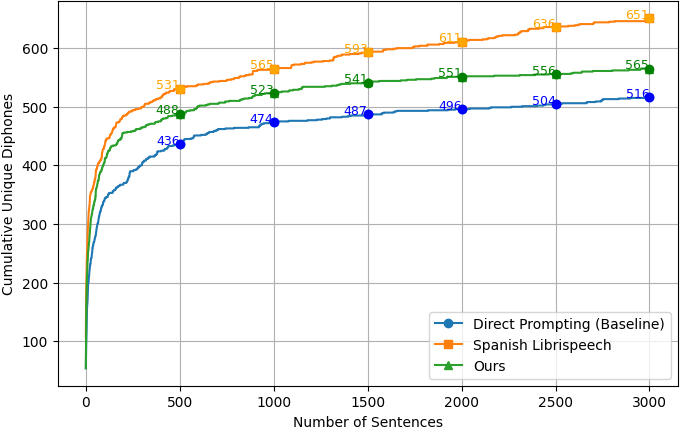}
  \caption {\label{fig:diphone-coverage}Diphone coverage for different dataset sizes for Baseline, Librispeech and Our Pipeline for English and Spanish text scripts}
\end{figure*}
\begin{figure*}[htbp]
  \includegraphics[width=1.0\linewidth]{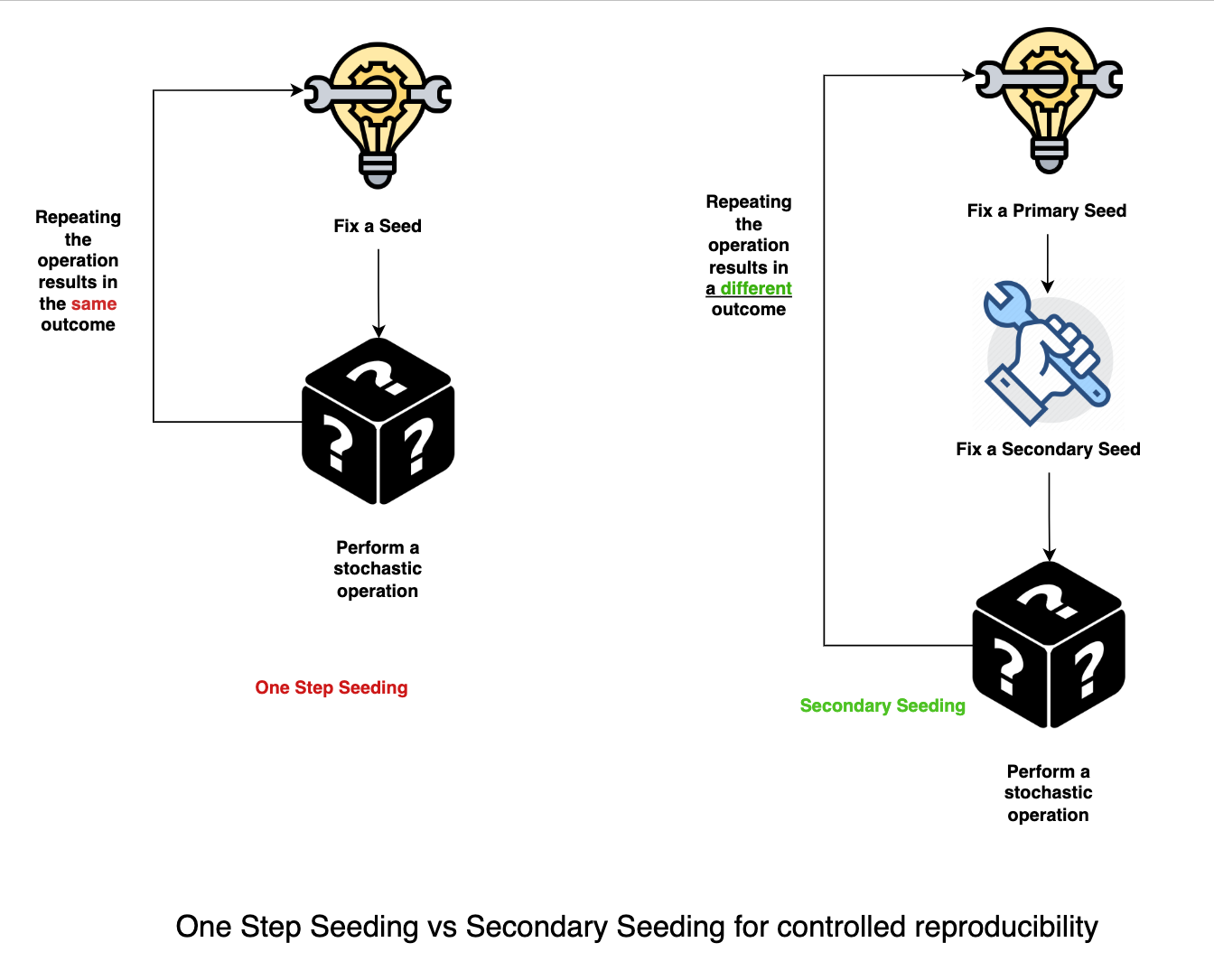}
  \caption {\label{fig:secondary-seed}Detailed description of text script generation pipeline}
\end{figure*}

\end{document}